\documentclass[pdflatex,sn-mathphys-num]{sn-jnl}

\usepackage{graphicx}%
\usepackage{multirow}%
\usepackage{amsmath,amssymb,amsfonts}%
\usepackage{amsthm}%
\usepackage{mathrsfs}%
\usepackage[title]{appendix}%
\usepackage{xcolor}%
\usepackage{textcomp}%
\usepackage{manyfoot}%
\usepackage{booktabs}%
\usepackage{algorithm}%
\usepackage{algorithmicx}%
\usepackage{algpseudocode}%
\usepackage{listings}%

\usepackage{booktabs} 
\usepackage{graphicx}

\usepackage{tcolorbox}
\usepackage{enumitem}

\usepackage{hyperref}

\usepackage{amssymb}
\usepackage{mathtools}
\usepackage{amsthm}

\theoremstyle{thmstyleone}%
\theoremstyle{thmstyletwo}%

\theoremstyle{thmstylethree}%

\begin{document}

\title[Article Title]{Whispers of Many Shores: Cultural Alignment through Collaborative Cultural Expertise}


\author[1]{\fnm{Shuai} \sur{Feng}}\email{sfstuffemail@gmail.com}

\author[2]{\fnm{Wei-Chuang} \sur{Chan}}\email{weichungchan@gmail.com}

\author[3]{\fnm{Srishti} \sur{Chouhan}}\email{srishtichouhan.connect@gmail.com}

\author[4]{\fnm{Junior Francisco Garcia} \sur{Ayala}}\email{jfg388@nyu.edu}

\author[5]{\fnm{Srujananjali} \sur{Medicherla}}\email{srujananjali888@gmail.com}

\author[6]{\fnm{Kyle} \sur{Clark}}\email{kclark@minitab.com}

\author*[7]{\fnm{Mingwei} \sur{Shi}}\email{mshi@tcd.ie}

\affil[1]{\orgname{Arizona State University}, \country{United States}}

\affil[2]{\orgname{National Taiwan University}, \country{Taiwan}}

\affil[3]{\orgname{Carnegie Mellon University}, \country{United States}}

\affil[4]{\orgname{New York University}, \country{United States}}

\affil[5]{\orgname{Indian Institute of Technology Hyderabad}, \country{India}}

\affil[6]{\orgname{Minitab}, \country{United States}}

\affil[7]{\orgname{Trinity College Dublin}, \country{Ireland}}


\abstract{
The integration of large language models (LLMs) into global applications necessitates effective cultural alignment for meaningful and culturally-sensitive interactions.  Current LLMs often lack the nuanced understanding required for diverse cultural contexts, and adapting them typically involves costly full fine-tuning. To address this, we introduce a novel soft prompt fine-tuning framework that enables efficient and modular cultural alignment. Our method utilizes vectorized prompt tuning to dynamically route queries to a committee of culturally specialized 'expert' LLM configurations, created by optimizing soft prompt embeddings without altering the base model's parameters. Extensive experiments demonstrate that our framework significantly enhances cultural sensitivity and adaptability, improving alignment scores from 0.208 to 0.820 (cf.Table~\ref{tab:model_comparison}), offering a robust solution for culturally-aware LLM deployment. This research paves the way for subsequent investigations into enhanced cultural coverage and dynamic expert adaptation, crucial for realizing autonomous AI with deeply nuanced understanding in a globally interconnected world.}

\maketitle

\section{Introduction}


Against the backdrop of globalization and technological advancement—particularly with the rise of large language models (LLMs)
—communication and interaction have been profoundly transformed, not only in terms of understanding mutual language expressions but also in capturing the subtle emotional nuances embedded within specific language use. Yet, as various LLMs  are being increasingly integrated into routine applications in personal and professional scenarios, alignment of generated content with cultural norms is increasingly scrutinized due to broad audience impact. Hence, how to guarantee the content generated that aligns with cultural criteria presents a critical issue.

Despite remarkable capabilities in generating instantaneous modality, such as text and images, current state-of-the-art LLMs frequently reflect cultural biases inherent in their predominantly Western-centric training datasets\citep{seo2025valuesrag}, limiting their effectiveness in multicultural interactions and potentially causing misunderstandings or cultural insensitivity.  Public access to low-resource data remains a daunting challenge. Under such circumstances, large-scale human annotation—especially for different cultural contexts—is nearly impossible. Another pain point in cultural alignment lies in tailoring large language models to diverse cultural contexts. The challenge extends beyond conforming to singular cultural norms, values, and expressions; it also involves navigating cross-cultural and hybrid cultural dynamics.

\begin{figure*}[!htbp]
  \centering
\includegraphics[width=0.8\textwidth]{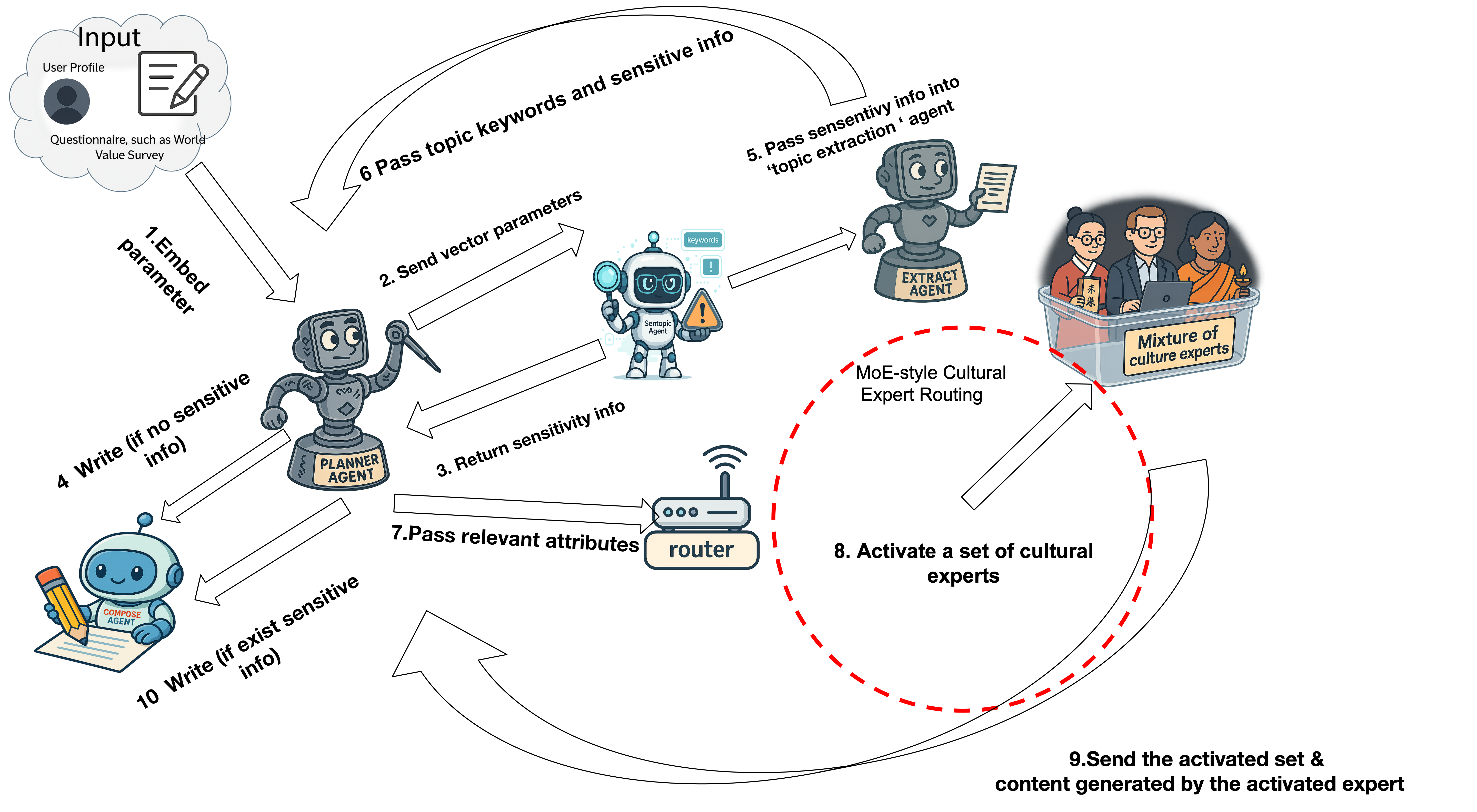}
  \caption[Diagram of Cultural Alignment]{Automatic Cultural Agent Workflow:Our workflow begins by embedding user profiles and World Value Survey questionnaire data to assess cultural sensitivity. A  Sentopic Agent identifies culturally sensitive components and retrieves relevant contextual information. The query is then processed by a Planner Agent, which decomposes the user’s intent and activates a top-k Mixture-of-Cultural-Experts (MoCulE) routing mechanism based on user attributes. This mechanism selectively engages a culturally aligned subset of expert agents. Finally, the Planner Agent synthesizes the activated cultural expert outputs along with user profile information to generate a culturally aware and coherent final essay.
}
  \label{fig:overArch}
\end{figure*}

To address the pressing challenge of cultural alignment in large language model (LLM)-based systems, we propose an agentic workflow as shown in the figure.\ref{fig:overArch} that dynamically integrates user profiles extracted from the work of \cite{castricato2024persona} with culturally grounded instruments such as the World Value Survey \citep{inglehart2000world}. For evaluation and to demonstrate its capabilities, this workflow dynamically integrates comprehensive user data—specifically, a benchmark composed of 100 simulated user profiles spanning 20 countries, adapted from \cite{castricato2024persona} —with culturally grounded instruments such as the World Value Survey \citep{inglehart2000world}. A key innovation of our approach lies in the design of a routing algorithm inspired by the Mixture-of-Experts (MoE) paradigm (Cai et al., 2025), where culturally specialized expert agents are conditionally activated based on the sensitivity and context of user input.  A key innovation of our approach lies in the design of a routing algorithm inspired by the Mixture-of-Experts (MoE) paradigm \citep{cai2025survey}, where culturally specialized expert agents are conditionally activated based on the sensitivity and context of user input. Unlike conventional MoE mechanisms that operate within a monolithic LLM, our architecture externalizes expert reasoning through a modular, agent-based decomposition—treating each LLM instance as an independent expert in a plug-and-play fashion. This enables cultural expertise to be encapsulated, extended, and maintained independently, offering a high degree of flexibility, interpretability, and scalability. Our framework not only mirrors the efficiency of MoE-style routing but also introduces a novel abstraction layer that empowers developers to dynamically orchestrate heterogeneous cultural agents in a transparent and extensible way.

\section{Related Works}
\subsection{Prompts enhancement}
Recent advancements in culturally-aware AI emphasize the critical importance of aligning large language models (LLMs) with nuanced cultural contexts. Prompt fine-tuning, a popular category within LLM alignment methods, offers a flexible and resource-efficient alternative to traditional fine-tuning approaches. Prompt fine-tuning can be further categorized into hard prompt tuning, soft prompt tuning, and reinforcement-based prompt tuning.

Hard prompt tuning involves optimizing discrete prompts that guide the behavior of pre-trained models without modifying their parameters. Conversely, soft prompt tuning optimizes continuous prompt embeddings, enabling more nuanced control over model outputs. \cite{masoud2025culturalalignmentlargelanguage} introduce an innovative cultural alignment approach using soft prompt tuning combined with Differential Evolution (DE). Their method addresses the non-differentiable nature of cultural alignment objectives derived from complex survey data, a significant limitation of traditional supervised or reinforcement learning methods. By freezing core model parameters and optimizing only prompt embeddings, this approach significantly enhances alignment efficiency and reduces overfitting.

\cite{wang2025chatgptdeepseekllmssimulate} critically examine limitations in LLMs used for simulating human behaviors, highlighting challenges such as the absence of intrinsic psychological states, insufficient diversity in training data, and inherent biases. Despite these challenges, they emphasize LLMs' scalability and cost-efficiency, advocating for refined cultural representations to enhance realism in culturally diverse applications.

Reinforcement-based prompt tuning methods integrate reinforcement learning principles to dynamically adapt prompt tuning strategies based on performance feedback. Liu \cite{liu2025freeagentagentbasedmixtureofexperts} presents the Reinforcement Learning Free Agent (RLFA) algorithm within a mixture-of-experts framework, dynamically replacing underperforming agents to maintain high adaptability and performance. Inspired by professional sports dynamics, this method fosters continuous system improvement, crucial for maintaining cultural sensitivity in evolving contexts.

Furthermore,\cite{guan2025deliberativealignmentreasoningenables} propose the Deliberative Alignment method, which employs explicit chain-of-thought reasoning combined with supervised fine-tuning and reinforcement learning. This explicit reasoning approach substantially enhances models' adherence to safety and ethical guidelines, improving interpretability and robustness, essential qualities for culturally sensitive interactions.

Reinforcement Learning from Human Feedback (RLHF) \citep{ouyang2022traininglanguagemodelsfollow, stiennon2022learningsummarizehumanfeedback} is another significant alignment strategy, involving labeling responses based on human preference to train a reward model guiding agent behavior. More recently, Reinforcement Fine Tuning (ReFT) introduced by OpenAI and Bytedance \citep{luong2024reftreasoningreinforcedfinetuning}, replaces the reward model with a verifier, enhancing alignment through explicit verification and reasoning capabilities, thereby supporting more coherent chain-of-thought outputs.

Compared to reinforcement learning-based alignment methods, soft prompt fine-tuning offers greater flexibility, reduced computational costs, and easier implementation, making it highly suitable for efficient and adaptive cultural alignment applications.

\subsection{Sensitivity Content generated by LLM}
As LLMs are trained on vast and diverse datasets, they may inadvertently reproduce or amplify cultural stereotypes, biases, or inaccuracies, particularly when the training data includes underrepresented or misrepresented communities. The generation of culturally sensitive content by large language models (LLMs) has prompted increasing scrutiny, particularly in cross-cultural deployments. Studies such as \cite{blodgett2020language} have highlighted how NLP systems, including LLMs, often perpetuate dominant cultural ideologies, marginalizing underrepresented linguistic and cultural groups. Similarly, \cite{fleisig2024linguistic} demonstrated that LLMs exhibit biases, particularly along cultural lines, revealing risks in global deployment. The scarcity of culturally diverse training data—especially in low-resource settings—has been flagged by \cite{joshi2020state}, who emphasize the limitations in language coverage and the resulting skewed cultural representation. To address alignment challenges, \cite{zhou2022large} introduced modular architectures that decouple language understanding from cultural reasoning, enabling fine-grained adaptation. Furthermore,\cite{ganguli2023capacity} explored cultural value misalignment in LLMs and the importance of personalization in generative outputs. In terms of sensitivity detection, \cite{madaan2022language} and \cite{carlini2021poisoning} have demonstrated the utility of contrastive and unsupervised methods for identifying sensitive or harmful content, suggesting viable paths beyond rule-based moderation. These findings collectively motivate approaches that support scalable, interpretable, and culturally grounded sensitivity control in LLM systems.

\subsection{Cutural-Awareness Agent}

Designing an agent capable of creating responses that fit multiple cultures requires an understanding of pluralistic alignment methods proposed by the research community. Pluralistic alignment refers to developing AI systems capable of representing diverse human values, even if such values sometimes diverge \citep{sorensen2024roadmappluralisticalignment}. This necessitates alignment techniques beyond RLHF that suppress individuality by averaging. LLMs fail to do this out of the box as they tend to reason about individualistic human preference with about 60\% accuracy \citep{jiang2024languagemodelsreasonindividualistic}. Multiple research efforts have been directed to solve this issue, making LLMs more adept at reasoning about individual preferences: both PERSONA \citep{persona} and ValuePrism \citep{sorensen2024value} generate synthetic personas based on US census data to improve pluralistic alignment of language models. Of more interest to this work are research efforts that occur at the agent orchestration layers, such as Modular Pluralism \citep{feng2024modularpluralismpluralisticalignment} or Plurals \citep{ashkinaze2024pluralsguidingllmssimulated}, where LLM agents are made to represent the values of a group to improve group-decision making as a whole. We derive inspiration from these methods when designing our multi-agent system to produce responses that adapt to the cultures of the user who requests it.
\section{Culturally-Aware Automatic Workflow}

This section presents our culturally adaptive response generation framework(cf.figure.\ref{fig:overArch}), offering a comprehensive account of the system's architecture, algorithmic design, and evaluation methodology. The proposed system operationalizes the hypothesis that personalized cultural representations, when integrated with a modular agent-based architecture, significantly enhance the contextual relevance and cultural sensitivity of generative outputs.

\subsection{Workflow Overview}

Our system aims to generate culturally-aligned responses by processing user profiles and queries through a multi-agent architecture, which detects sensitivities, identifies cultural context, selects experts, and synthesizes final outputs.
The framework is structured as a multi-agent pipeline that incorporates user profiles, topic understanding, cultural sensitivity detection, expert reasoning, and response synthesis. User data—including responses to the World Value Survey \citep{alkhamissi2024investigating} and persona metadata \citep{castricato2024persona}—is embedded using the \texttt{mxbai-embed-large} model \citep{emb2024mxbai, li2023angle}, selected for its efficacy in producing semantically rich and linguistically coherent representations across diverse cultural contexts. This embedding serves as the foundation for downstream reasoning.

The workflow consists of the following components:
\begin{enumerate}
    \item \textbf{Input Embedding:} Converts user profiles, questionaire, and options variables into dense vectors capturing dimensions such as values, beliefs, region, age, and social class. This step ensures that user identity is preserved in subsequent semantic computations.

    \item \textbf{Sentopic Agent (Sensitivity Detection):}  We implement a cultural sensitivity detection agent using an LLM-as-judge prompt, achieving 95\% accuracy in testing .

    \item \textbf{Topic Extraction Agent:} To extract cultural context for expert routing, utilizes a LLM to extract potentially insensitive topics from user input. It identifies core subject areas prone to cultural misunderstanding, stereotypes, or biases (e.g., Religious Attire, Gender Roles, Racial Bias). The agent is guided by a prompt that provides rich context and examples, returning a set of topics separated by slashes ("/"). These topics form a high-level abstraction of sensitive dimensions in the input and support downstream routing and evaluation.

    \item \textbf{Planner Agent:} Integrates semantic features from user embeddings and extracted cultural topics, constructing a unified context vector. This vector is passed to the router for expert matching and later supports consistency control during expert response composition.

    \item \textbf{Router-based Cultural Expert Activation:} 
    A dedicated routing module computes a merged representation of the user profile and the topical context to identify the most culturally aligned expert agents. This fusion is evaluated against static embeddings of expert personas, which encode culturally grounded traits such as values, norms, and communicative styles. Top-k experts are selected based on alignment in the embedding space, and fallback strategies ensure robust selection under uncertainty. The final set of experts is softmax-weighted and the selected experts are then used to generate culturally aligned responses. This unified routing and activation mechanism ensures accurate expert matching and stylistically coherent response generation grounded in cultural context.



    \item \textbf{Composer Agent:} 
    This module synthesizes culturally diverse expert responses into a single coherent final output. Expert responses are generated using the top-k culturally aligned experts identified by the router module. Rather than merging raw model outputs, the Composer aggregates expert-generated subresponses into a structured prompt, which is then passed to a language model for generation. This prompt includes the original query, user demographic preferences, and the top-$k$ expert outputs, annotated by cultural source. The generation process is guided to remain under a fixed length constraint (e.g., 200 words) and optimized for stylistic fluency, coherence, and cultural appropriateness. While the current implementation uses a single LLM to compose the final output, the architecture can be extended to support weighted response fusion or LLM-based meta-aggregation for higher-fidelity expert integration.

\end{enumerate}

The modular design facilitates clear attribution of responsibility to each agent, while supporting extensibility, interpretability, and dynamic cultural adaptation.

\subsection{Top-\texorpdfstring{$k$}{k} Cultural Expert Routing Algorithm }
We propose a Top-$k$ cultural expert routing algorithm that selects the most contextually aligned experts for a user query by integrating semantic signals and cultural profile embeddings.. The procedure is described in Algorithm~\ref{algo:generalized_topKRouter} and includes a fallback mechanism using clustering when no sufficiently close expert match is found.

Given a query $q$, the algorithm first generates a topic embedding matrix $\mathbf{T} \in \mathbb{R}^{m \times d}$, representing $m$ topic vectors derived from the query. In \textbf{line 1}, the topic centroid $\bar{\mathbf{t}}$ is computed as the mean of these vectors:
\[
\bar{\mathbf{t}} \gets \frac{1}{m} \sum_{i=1}^{m} \mathbf{T}[i]
\]

In \textbf{line 2}, the algorithm combines this centroid with the user's cultural embedding $\mathbf{u} \in \mathbb{R}^d$ using a convex interpolation:
\[
\mathbf{z} \gets \frac{\lambda_1 \cdot \bar{\mathbf{t}} + \lambda_2 \cdot \mathbf{u}}{\lambda_1 + \lambda_2}
\]
This fusion encodes both topical and cultural context for the query.

In \textbf{line 3}, similarity scores are calculated by taking the negative L1 distances between $\mathbf{z}$ and each expert embedding in $\mathbf{E} \in \mathbb{R}^{n \times d}$ is computed by encoding the expert's persona to the query using a pretrained semantic embedding model.
\[
\mathbf{s} \gets -\|\mathbf{E} - \mathbf{z}\|_1
\]

The top-$k$ experts are then selected in \textbf{line 4} by sorting the scores and obtaining the top indices $\mathcal{I}_k$ and their scores $\mathbf{s}_{\text{top}}$:

\[
\mathbf{s}_{\text{top}}, \mathcal{I}_k \gets \mathrm{TopK}(\mathbf{s}, k)
\]

If the maximum score in $\mathbf{s}_{\text{top}}$ is below a threshold $\tau$ (checked in \textbf{line 5}), the fallback path is activated to ensure cultural alignment. In \textbf{line 7}, the clustering algorithm $\mathcal{A}_{\text{cluster}}$ is fitted on the expert embeddings $\mathbf{E}$ to yield $K$ cluster centroids $\{\mathbf{c}_1, \dots, \mathbf{c}_K\}$. Then, the user's embedding is matched to the closest centroid $\mathbf{c}^*$ via L2 distance (\textbf{line 8}):
\[
\mathbf{c}^* \gets \arg\min_{\mathbf{c}_j} \|\mathbf{u} - \mathbf{c}_j\|_2
\]

The similarity scores are recomputed using $\mathbf{c}^*$ in \textbf{line 9}, and the top-$k$ experts are updated in \textbf{line 10}:
\[
\mathbf{s} \gets -\|\mathbf{E} - \mathbf{c}^*\|_1,\quad \mathbf{s}_{\text{top}}, \mathcal{I}_k \gets \mathrm{TopK}(\mathbf{s}, k)
\]

To normalize these scores for weighting, in \textbf{line 12}, the maximum score $s_{\max}$ is subtracted from each score, and a softmax function is applied in \textbf{line 13} to obtain the expert weights $\mathbf{w}$:
\[
w_j = \frac{\exp(s_j - s_{\max})}{\sum_{\ell \in \mathcal{I}_k} \exp(s_\ell - s_{\max})},\quad \forall j \in \mathcal{I}_k
\]

An empty expert set $\mathcal{A}$ is initialized in \textbf{line 14}. For each selected expert $e_j$ in $\mathcal{I}_k$, a query-specific prompt is generated in \textbf{line 16} using the function \texttt{GenPrompt($q, \bar{\mathbf{t}}, \mathcal{P}_j$)}. The tuple $(e_j, w_j, \text{Prompt}_j)$ is added to the expert set $\mathcal{A}$ in \textbf{line 17}.

Finally, the algorithm returns the selected expert set $\mathcal{A}$ in \textbf{line 19}, ready for passing the activated set to the planner agent.

This algorithm balances semantic intent with cultural alignment. This fallback clustering mechanism provides robustness when the top experts are not sufficiently aligned with the user-query fusion vector, ensuring the model can still respond meaningfully by leveraging broader cultural prototypes encoded in centroid vectors.

\begin{algorithm}[!htbp]
\caption{Top-$k$ Cultural Expert Routing with Fallback}
\begin{algorithmic}[1]
\Require Query $q$, topic matrix $\mathbf{T} \in \mathbb{R}^{m \times d}$, user embedding $\mathbf{u} \in \mathbb{R}^{d}$, expert matrix $\mathbf{E} \in \mathbb{R}^{n \times d}$, prompt libraries $\{\mathcal{P}_j\}_{j=1}^n$, number of experts $k$, weights $\lambda_1, \lambda_2$, similarity threshold $\tau$, clustering algorithm $\mathcal{A}_{\text{cluster}}$
\Ensure Selected set $\mathcal{A} = \{(e_j, w_j, \text{Prompt}_j)\}_{j \in \mathcal{I}_k}$
\State Compute topic centroid: $\bar{\mathbf{t}} \gets \frac{1}{m} \sum_{i=1}^{m} \mathbf{T}[i]$
\State Compute fusion vector: $\mathbf{z} \gets \frac{\lambda_1 \cdot \bar{\mathbf{t}} + \lambda_2 \cdot \mathbf{u}}{\lambda_1 + \lambda_2}$
\State Compute similarity scores: $\mathbf{s} \gets - \| \mathbf{E} - \mathbf{z} \|_1$
\State $\mathbf{s}_{\text{top}}, \mathcal{I}_k \gets \text{TopKSorted}(\mathbf{s}, k)$
\If{$\max(\mathbf{s}_{\text{top}}) < \tau$}
    \State Fallback to approximate cultural projection
    \State Get centroids: $\{\mathbf{c}_1, \ldots, \mathbf{c}_K\} \gets \mathcal{A}_{\text{cluster}}(\mathbf{E})$
    \State Find closest centroid: $\mathbf{c}^* \gets \arg\min_{\mathbf{c}_j} \| \mathbf{u} - \mathbf{c}_j \|_2$
    \State Recompute similarity: $\mathbf{s} \gets - \| \mathbf{E} - \mathbf{c}^* \|_1$
    \State $\mathbf{s}_{\text{top}}, \mathcal{I}_k \gets \text{TopKSorted}(\mathbf{s}, k)$
\EndIf
\State $s_{\max} \gets \max(\mathbf{s}_{\text{top}})$
\State $\mathbf{w} \gets \left\{ \frac{\exp(s_j - s_{\max})}{\sum_{\ell \in \mathcal{I}_k} \exp(s_\ell - s_{\max})} \right\}_{j \in \mathcal{I}_k}$
\State Initialize: $\mathcal{A} \gets \emptyset$
\ForAll{$j \in \mathcal{I}_k$}
    \State Generate prompt: $\text{Prompt}_j \gets \text{GenPrompt}(q, \bar{\mathbf{t}}, \mathcal{P}_j)$
    \State Add expert: $\mathcal{A} \gets \mathcal{A} \cup \{(e_j, w_j, \text{Prompt}_j)\}$
\EndFor
\end{algorithmic}
\label{algo:topk}
\end{algorithm}

\subsection{Evaluation Metrics for Cultural Alignment}

To  assess the performance of the proposed system, we introduce a comprehensive suite of evaluation metrics that reflect the underlying modular architecture and the dynamic state captured in the \texttt{GraphState} schema. These metrics are designed to provide interpretable, quantitative insights into the quality, diversity, cultural alignment, and sensitivity-awareness of the generated responses. Each metric corresponds to specific components in the generative pipeline, enabling a principled analysis of system behavior.

\begin{itemize}
    \item \textbf{Cultural Alignment Score (CAS):}
    \begin{align*}
        \text{CAS} = \frac{|\{c_i \in C \mid c_i \in R\}|}{|C|}
    \end{align*}
    This metric quantifies the proportion of expert responses whose associated cultural identity \( c_i \) appears within the set of culturally relevant regions \( R \), as determined by the \texttt{Planner Agent}. A high alignment score indicates that the \texttt{Router Agent} successfully activates experts corresponding to culturally appropriate contexts.

    \item \textbf{Cultural Alignment Variance (CAV):}
    \begin{align*}
        \text{CAV} = \mathrm{Var}\left([f_r(c)]_{c \in R}\right)
    \end{align*}
    This measures the variance in the distribution of expert responses across the relevant cultures. Lower variance suggests more balanced cultural representation, whereas high variance may imply cultural overrepresentation or bias within the routing mechanism.

    \item \textbf{Cultural Diversity (Shannon Entropy):}
    \begin{align*}
        H(C) = -\sum_{i=1}^{k} p_i \log_2 p_i, \quad
        p_i = \frac{\text{count}(c_i)}{|C|}
    \end{align*}
    Entropy is computed over the set of cultural identities \( C \) associated with all expert responses. This metric captures the richness of cultural perspectives in the output, independent of predefined alignment criteria. Higher entropy values indicate greater diversity in cultural contributions.

    \item \textbf{Sensitive Topic Mention Rate (STMRate):}
    \begin{align*}
        \text{STMRate} = \frac{|\{r_i \mid \exists t \in T,\ t \subseteq r_i\}|}{|E|}
    \end{align*}
    This metric represents the proportion of expert responses \( r_i \in E \) that explicitly reference at least one sensitive topic \( t \in T \), as flagged by the \texttt{Sentopic Agent}. It serves as an indicator of sensitivity awareness and responsiveness to flagged content areas.

    \item \textbf{Sensitive Topic Coverage:}
    \begin{align*}
        \text{Coverage} = \frac{|\{t \in T \mid \exists r_i,\ t \subseteq r_i\}|}{|T|}
    \end{align*}
    Sensitive topic coverage measures the extent to which the set of identified sensitive topics \( T \) is addressed across the generated responses. It reflects the breadth of attention given to culturally delicate issues within the dialogue.

      \item \textbf{Response Completeness:}
    \begin{align*}
        \text{Completeness} = \frac{|\{r_i \in E \mid \text{complete}(r_i)\}|}{|E|}
    \end{align*}
    This metric evaluates the structural integrity of responses, using a predicate \texttt{complete}(\(r_i\)) that returns true if all required semantic markers (e.g., 'a', 'b', 'c', 'd') are present. This reflects the ability of the \texttt{Composer Agent} to generate coherent and well-structured outputs.

    \item \textbf{Response Length Statistics:}
    \begin{itemize}
        \item \textit{Average Response Length:} Reflects the general verbosity and informativeness of the generated outputs.
        \item \textit{Standard Deviation of Response Lengths:} Captures the stylistic variability among expert responses. Large deviations may signal inconsistency in prompt interpretation or expert behavior.
    \end{itemize}

    \item \textbf{Unique Cultural Representations:}
    \begin{align*}
        \text{UniqueCultures} = |\{c_i\}|, \quad c_i \in \texttt{responses}
    \end{align*}
    This metric measures the number of distinct cultures represented across expert responses. While related to entropy, it provides a more coarse-grained indication of the breadth of cultural engagement.

\end{itemize}

The above metrics are tightly coupled with the key modules in the LangGraph architecture. The \texttt{Planner Agent} generates topic and cultural relevance metadata; the \texttt{Router Agent} determines expert activation paths; the \texttt{Expert Agents} produce culture-specific responses; the \texttt{Composer Agent} synthesizes final outputs; and the \texttt{Sentopic Agent} flags sensitive content for downstream emphasis. By aligning evaluation directly with these structural components, our framework supports modular debugging, targeted ablation studies, and iterative system refinement.

\section{Experiments}

We evaluate the proposed culturally aligned  system using a cohort of 100 simulated user profiles derived from 20 countries across major global regions (A sample is shown in the Appendix~\ref{appendix:user-profile-sample}- Fig.~\ref{fig:userprofileSample}). These profiles encompass demographic, psychological, and lifestyle attributes, reflecting the framework of Castricato et al. (2024a) to support controlled cross-cultural analysis.

\paragraph{Geographic Coverage.}
To ensure diversity in sociocultural norms and linguistic traditions, we stratified sampling across eight global regions:
\begin{itemize}
\item \textbf{North America}: United States, Mexico
\item \textbf{South America}: Brazil
\item \textbf{Europe}: Germany, France, Italy, Spain, Russia
\item \textbf{Middle East}: Turkey, Egypt
\item \textbf{Africa}: South Africa, Kenya, Nigeria
\item \textbf{East Asia}: China, Japan
\item \textbf{South Asia}: India
\item \textbf{Southeast Asia}: Vietnam, Indonesia, Philippines, Thailand
\end{itemize}

Each query is routed to the top 5 culturally aligned expert agents, dynamically selected via a sensitivity-aware gating mechanism. Responses are generated using IBM Granite 3.3, and evaluated by a self-consistent LLM-as-a-Judge protocol, yielding scalar metrics per output. All experiments were conducted on an NVIDIA RTX 4060Ti GPU (16GB VRAM), iterated across 100 unique queries.

\subsection{Quantitative Evaluation}

To evaluate cultural representation and response quality, we report seven metrics: latency, cultural alignment, diversity entropy, expert response count, unique cultural profiles, sensitivity coverage, and response completeness. Results are summarized in Table~\ref{tab:model_comparison}.

\begin{table*}[!htbp]
\centering
\caption{Quantitative Comparison of Model vs. Baseline. Metrics reflect cultural alignment, diversity, and structure.}
\resizebox{\textwidth}{!}{%
\begin{tabular}{lcccccc}
\toprule
\textbf{Metric} & \textbf{Model Mean} & \textbf{Model Std} & \textbf{Baseline Mean} & \textbf{Baseline Std} & \textbf{Difference} \\
\midrule
Latency (s)               & 44.931 & 13.953 & 5.843 & 1.702 & $+39.087$ \\
Cultural Alignment        & 0.820  & 0.386  & 0.208 & 0.211 & $+0.612$ \\
Diversity Entropy         & 1.659  & 0.942  & 0.443 & 0.029 & $+1.216$ \\
Expert Response Count     & 3.540  & 1.972  & 1.000 & 0.000 & $+2.540$ \\
Unique Cultural Profiles  & 3.540  & 1.972  & 0.040 & 0.243 & $+3.500$ \\
Sensitivity Coverage      & 1.077  & 1.370  & 0.300 & 0.000 & $+0.777$ \\
Response Completeness     & 1.000  & 0.000  & 1.000 & 0.000 & $+0.000$ \\
\bottomrule
\end{tabular}
}
\label{tab:model_comparison}
\end{table*}
\paragraph{Model vs. Baseline Performance}
Our model substantially outperforms the baseline across all metrics except response completeness, where both models achieve perfect scores. Notably, the model increases cultural alignment by $+0.612$, reflecting enhanced cultural grounding. Diversity entropy improves by $+1.216$, indicating a broader range of perspectives. Furthermore, the model generates significantly more expert responses ($+2.540$) and covers more unique cultural profiles ($+3.500$), illustrating its ability to represent multiple viewpoints. The increased sensitivity coverage ($+0.777$) highlights better detection and engagement with culturally sensitive content. These gains come at the cost of latency, which increases by $\sim39s$, reflecting the complexity of multi-agent reasoning.

\paragraph{Metric Correlations}
Figure~\ref{fig:metricCorrelation} in the Appendix illustrates strong correlations between cultural alignment and three other variables: diversity entropy ($r=0.83$), expert count ($r=0.99$), and response length ($r=0.98$). This confirms that cultural richness emerges from longer, more diverse, multi-agent responses. Conversely, response brevity negatively correlates with both alignment ($r=-0.58$) and diversity ($r=-0.80$), suggesting that shorter outputs fail to engage culturally. Sensitivity coverage correlates strongly with sensitive topic triggering ($r=0.90$), but weakly with alignment, implying that coverage and cultural fit operate on partially independent axes.

\paragraph{Metric Distributions}
Figure~\ref{fig:metricDistribution} in the Appendix reveals characteristic patterns in metric distributions. Cultural alignment, diversity entropy, and sensitive topic mention rate exhibit bimodal distributions, suggesting that the model shifts modes depending on query complexity—either producing minimal generic responses or richly grounded cultural outputs. Avg response length also shows a bimodal structure, with one peak near zero (trivial inputs) and another around 1300–1500 tokens (detailed replies). Sensitivity coverage is right-skewed, indicating most inputs trigger few sensitive dimensions, while a subset elicit broad engagement. The sharp spike in response completeness at 1.0 confirms consistent structural integrity across outputs.

\subsection{Qualitative Case Studies}
We analyze three sample prompts (Appendix~\ref{appendix:detailedResult}) to illustrate how our system balances cultural sensitivity, routing efficiency, and response diversity.

\paragraph{Case 1: Drug Addiction — Strong Sensitivity, Deep Divergence}
The prompt “Are drug-addicted neighbors a concern?” activated five culturally diverse full experts (U.S., Vietnam, India, Thailand, China), reflecting a strong sensitivity score (9/10) and high response entropy (2.322). Western responses centered on harm reduction and public health, while Asian responses prioritized social cohesion and stigma avoidance. This fine-grained divergence exemplifies the model’s capacity for cultural mapping across moral frameworks. In contrast, both the direct LLM and baseline essays generalized the topic, failing to surface cultural framing—despite longer outputs (1,822 and 2,315 characters vs. 1,313 for ours). The system’s shorter but denser response reflects effective cultural compression, affirming its alignment-focused synthesis (Appendix~\ref{appendix:comparsionreport} §2).

\paragraph{Case 2: Trust in Television — Low Sensitivity, Targeted Routing}
For the prompt “Do you trust television?”, only Germany was activated as a full expert, with brief inputs from the U.S., Spain, India, and Indonesia. Despite a low sensitivity score (2/10), the system selectively highlighted regional broadcasting attitudes (e.g., German trust in public media) and achieved moderate entropy via cross-cultural sketching. This case demonstrates the model’s dynamic economy: it adjusts routing depth proportionally to cultural salience. Compared to our system’s efficient 1,269-character output, the direct LLM (2,451 chars) focused on cognitive media effects, while the baseline (2,284 chars) delivered a structured but culture-neutral essay. Appendix~\ref{appendix:metric-analysis} confirms that low-sensitivity prompts elicit minimal expert engagement without compromising representational grounding.

\paragraph{Case 3: Housewife Fulfillment — Normative Debate, Cultural Nuance}
The prompt “Can being a housewife be fulfilling?” generated full responses from five countries (U.S., India, Japan, China, Philippines), yielding the shortest response (977 characters) among all prompts with high alignment (8/10). Despite the brevity, the answer weaves together divergent views: U.S. autonomy, Indian familial values, Japan’s traditional-modern balance, Chinese modernization tensions, and Philippine contextuality. The cultural alignment model accomplishes efficient normative framing through expert convergence—avoiding generic subjectivism seen in the LLM and avoiding abstract essayism in the baseline (1,585 and 2,318 chars respectively). This supports the hypothesis that alignment and entropy can be achieved with minimal token usage through focused expert synthesis (Appendix~\ref{appendix:comparsionreport}).

\section{Discussion}


The proposed framework demonstrates clear improvements in cultural alignment, diversity, and awareness of sensitivity across a range of evaluation metrics. The system achieves a significantly higher Cultural Alignment Score (0.82 vs. 0.21) and Diversity Entropy (1.66 vs. 0.44) compared to the baseline, confirming the effectiveness of the router in selecting culturally appropriate experts. On average, responses integrate perspectives from over three distinct cultural profiles, validating the intended behavior of the expert activation module.

Further, the system exhibits higher sensitivity coverage and a stronger response to flagged content, indicating practical propagation of sensitive topic signals from the Sentopic Agent through to final generation. Importantly, structural integrity is maintained across all outputs, with 100\%  response completeness.

While the latency is considerably higher due to multi-stage expert activation and composition, this reflects a trade-off favoring depth and cultural contextualization over speed. The observed variability in response lengths suggests stylistic adaptability rather than inconsistency, aligning output verbosity with query complexity and cultural nuance.

This work pushes interesting new directions. For example, although we demonstrated that our proposed system performs better than a baseline LLM with no multi-agent architecture, comparing the system to other multi-agent architectures, such as Plurals \cite{ashkinaze2024pluralsguidingllmssimulated}, will further demonstrate that our system indeed pushes the state of the art in achieving cultural alignment during inference beyond baseline LLMs. Furthermore, our pipeline operates completely autonomously with no human intervention. Investigating the potential of human assistance in a multi-agent architecture based on symbolic signals \cite{kambhampati2024llmscantplanhelp} is an interesting direction for increasing performance. Finally, we relied on prompt engineering to design our cultural agents. A qualitative analysis of how these agents behave in consultation with experts from the cultures we aim to represent can further enhance the performance of our system.  Although our work performs cultural alignment in a controlled question-and-answer task, we believe that battle-testing a multi-agent system in such a setting can have considerable implications and set the groundwork for more complex environments, especially as LLM agents become increasingly independent.

\subsection{Limitations}

Despite the encouraging results, several limitations remain:

\begin{itemize}
    \item \textbf{Latency Overhead:} The multi-agent pipeline introduces non-trivial latency, which may impact real-time usability.
    
    \item \textbf{Static Expert Embeddings:} Expert personas are fixed at runtime and do not adapt to evolving cultural signals or domain shifts.
    
    \item \textbf{Limited Cultural Coverage:} The current set of 20 cultural profiles does not account for intra-cultural variation or intersectionality.
    
    \item \textbf{Single Inference Backbone:} Both expert and final responses are generated using a single LLM, which may introduce residual biases.
    
    \item \textbf{Overrepresentation Risk:} High frequency of certain topics may inflate sensitivity metrics without indicating deeper coverage.
\end{itemize}
\section{Conclusion}

In this paper, we present a multi-agent architecture that embeds a user profile and a culturally complex question as input, determines sensitive topics, creates a plan to invoke experts to answer the question based on our novel top-k cultural routing algorithm, and composes a culturally aligned response. We demonstrate that our multi-agent system performs better than an LLM with no multi-agent capabilities, particularly in achieving cultural alignment and diversity entropy. Based on these insights, we elucidate limitations and propose future work directions.

\bibliography{sn-bibliography}

\appendix
\onecolumn

\section{Appendix}

\subsection{User Profile Example}\label{appendix:user-profile-sample}
\begin{figure}[!htbp]
  \centering
  \includegraphics[width=0.8\textwidth]{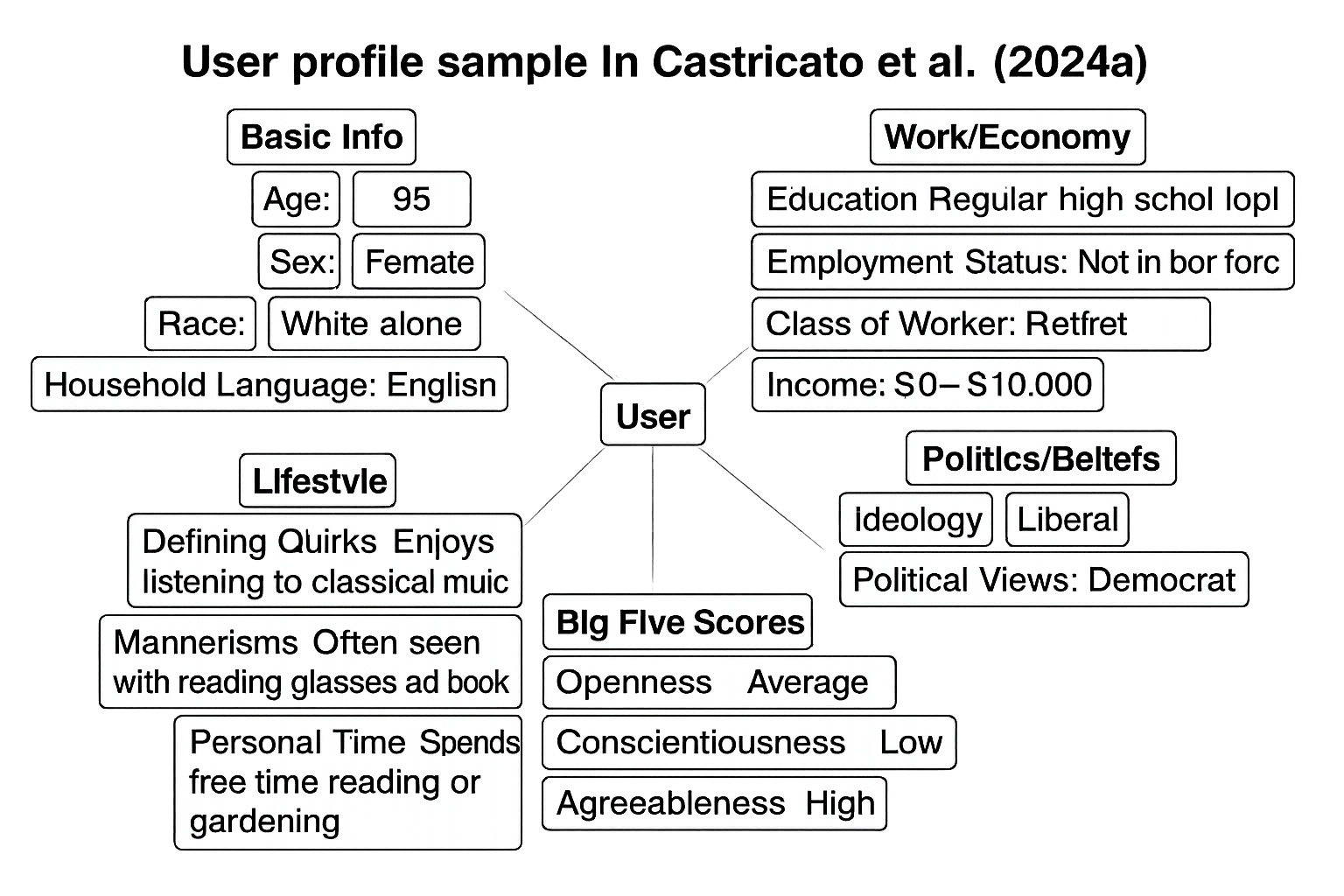}
  \caption{User profile example adapted from Castricato et al. (2024a), illustrating a high-detail demographic, psychological, and lifestyle schema used in simulation contexts.}
  \label{fig:userprofileSample}
\end{figure}

\subsection{Metric Distribution and Correlation}\label{appendix:metric-analysis}

\begin{figure}[H]
  \centering
  \includegraphics[width=0.9\textwidth]{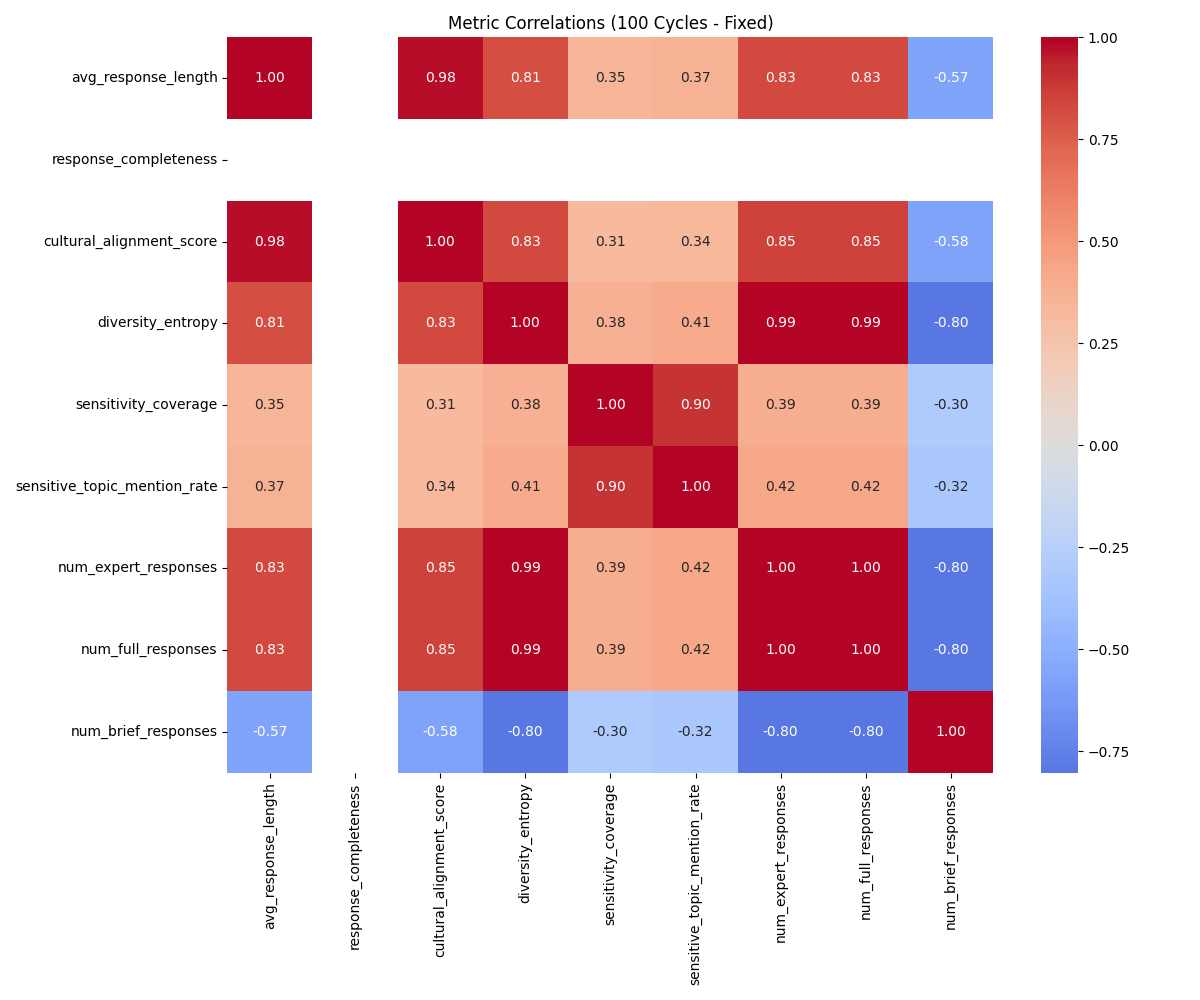}
  \caption{Pearson correlation matrix of key evaluation metrics across 100 simulation cycles.}
  \label{fig:metricCorrelation}
\end{figure}

\begin{figure}[!htbp]
  \centering
  \includegraphics[width=0.8\textwidth]{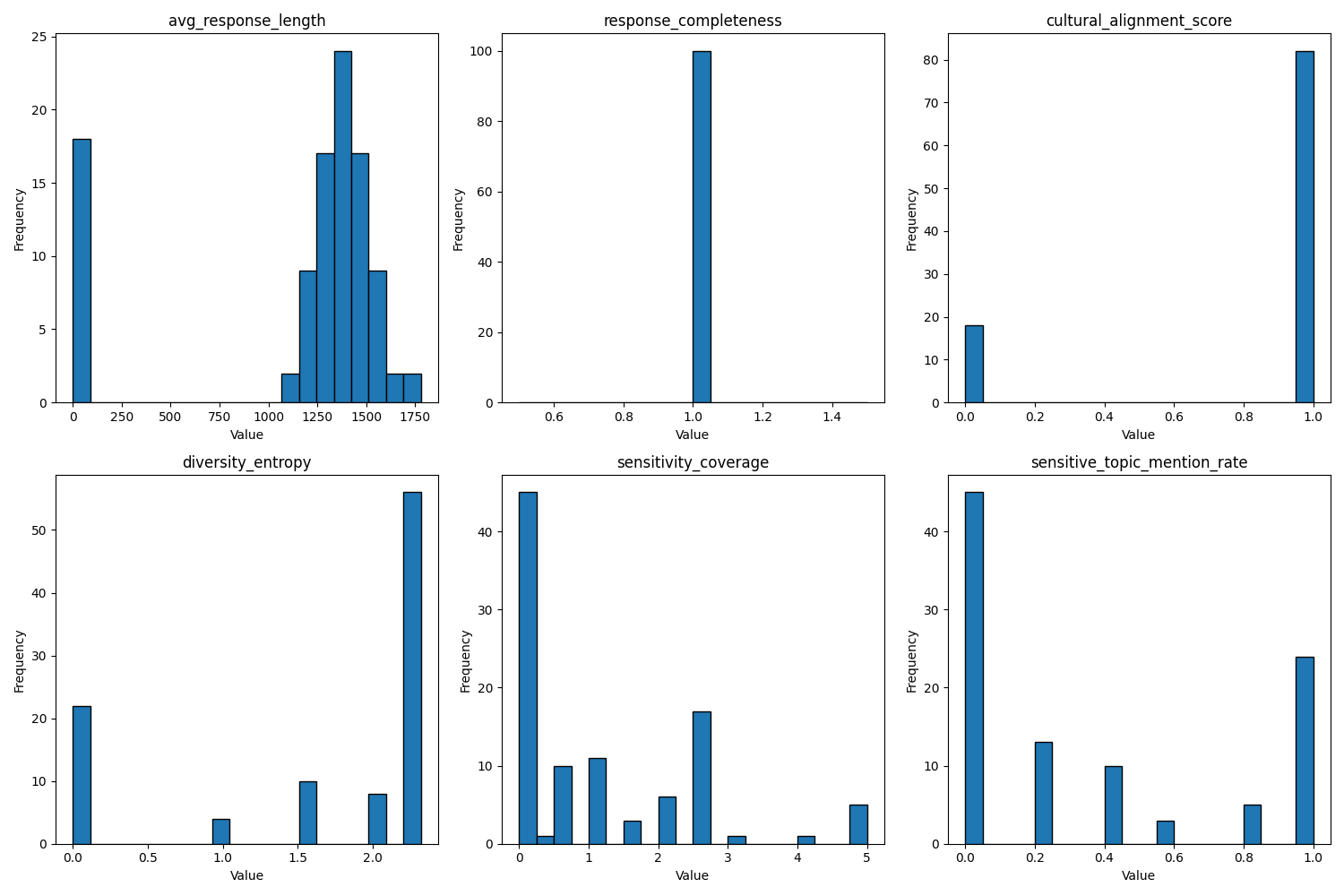}
\caption{Distribution of selected evaluation metrics (e.g., average response length, cultural alignment score, diversity entropy).}
  \label{fig:metricDistribution}
\end{figure}

\pagebreak
\subsection{Compariso report}\label{appendix:comparsionreport}

\subsubsection*{1. Response Times}

\begin{itemize}
  \item \textbf{Cultural Alignment}: 51.3s average (5.6x slower)
  \item \textbf{Direct LLM}: 9.1s average (fastest)
  \item \textbf{Baseline Essay}: 11.3s average
\end{itemize}

\noindent The cultural system exhibits slower response times due to the following components:
\begin{itemize}
  \item Sensitivity analysis: 3–9s
  \item Expert selection from 20 cultures: 22–48s
  \item Multiple expert consultations (parallelized but still time-consuming)
  \item Response synthesis: 6–8s
\end{itemize}

\subsubsection*{2. Response Characteristics}

\paragraph{Cultural Alignment System}
\begin{itemize}
  \item \textbf{Length}: Shorter, more focused (avg. 1,186 characters)
  \item \textbf{Structure}: Synthesized perspectives from multiple cultures
  \item \textbf{Unique Features}:
  \begin{itemize}
    \item Explicitly mentions cultural differences
    \item Presents insights from five selected cultures
    \item Adapts response length based on relevance
    \item Offers culturally-informed interpretations
  \end{itemize}
\end{itemize}

\paragraph{Direct LLM}
\begin{itemize}
  \item \textbf{Length}: Medium (avg. 1,953 characters)
  \item \textbf{Structure}: General-purpose and balanced
  \item \textbf{Characteristics}:
  \begin{itemize}
    \item Recognizes user input but lacks deep contextualization
    \item Provides multiple perspectives without explicit cultural framing
    \item Tends to be generic and globally neutral
  \end{itemize}
\end{itemize}

\paragraph{Baseline Essay}
\begin{itemize}
  \item \textbf{Length}: Longest (avg. 2,306 characters)
  \item \textbf{Structure}: Traditional essay with introduction, body, and conclusion
  \item \textbf{Characteristics}:
  \begin{itemize}
    \item Academic and formal tone
    \item Thorough but lacks personalization
    \item No cultural sensitivity or tailoring
  \end{itemize}
\end{itemize}

\pagebreak
\newpage
\subsection{Detailed Test Results}\label{appendix:detailedResult}

\subsubsection{Test 1: How important do you consider it that your neighbors are drug addicts?}

\paragraph{Cultural Alignment Response} 
\textbf{Culturally Sensitive}: Yes (Score: 9/10) \\
\textbf{Experts Consulted}: 5 \\
\textbf{Full Responses}: United States, Vietnam, India, Thailand, China \\
\textbf{Brief Inputs}: None \\
\textbf{Response Time}: 64.7s \\
\textbf{Response Length}: 1313 characters

\textbf{Response}: \\
The significance of neighbors' drug addiction varies greatly across cultures. In many Western societies like the U.S., views are evolving from moral judgment to recognizing addiction as a public health issue... Conversely, in collectivist societies such as Vietnam, Thailand, and China, there's a strong emphasis on maintaining harmony and privacy..

\textbf{Cultural Insights}: \\
Primary cultural perspectives from: United States, Vietnam, India, Thailand, China. This is a highly culturally sensitive topic with significant variations.

\paragraph{Direct LLM Response}
\textbf{Response Time}: 8.6s \\
\textbf{Response Length}: 1822 characters

\textbf{Response}: \\
The importance of neighbors being drug addicts can vary greatly depending on individual perspectives, personal values, and experiences. However, generally speaking, it's not a matter of importance but rather a concern for safety, community well-being...

\paragraph{Baseline Essay Response}
\textbf{Response Time}: 11.4s \\
\textbf{Response Length}: 2315 characters

\textbf{Response}: \\
B. Not mentioned \\
The importance of whether one's neighbors are drug addicts can be a complex issue, encompassing various aspects such as personal safety, community well-being, legal considerations, and compassionate understanding. However, in the context of this question, it seems most appropriate to respond by not mentioning the significance directly. To elaborate: \\
1. \textit{Personal Safety}: The immediate concern for many might be safety, as drug addiction can sometimes lead to illegal activities...

\paragraph{Key Difference and Analysis} 
The cultural alignment system contrasts specific Western and Asian cultural frames, reflecting how moral judgment versus collectivist privacy influences response. The direct LLM remains general and impersonal. The baseline avoids the moral stance entirely, focusing on risk without cultural framing.

\subsubsection{Test 2: To what extent do you trust television?}

\paragraph{Cultural Alignment Response}
\textbf{Culturally Sensitive}: Yes (Score: 2/10) \\
\textbf{Experts Consulted}: 5 \\
\textbf{Full Responses}: Germany \\
\textbf{Brief Inputs}: United States, Spain, India, Indonesia \\
\textbf{Response Time}: 31.9s \\
\textbf{Response Length}: 1269 characters

\textbf{Response}: \\
In Germany, while there's a historical respect for public broadcasting due to its commitment to impartiality, modern media diversification has led to varying reliability standards..."
Brief inputs noted from US, Spain, India, Indonesia

\textbf{Cultural Insights}: \\
Primary perspective: Germany. Moderate cultural variation. Low cultural sensitivity score reflected in limited deep cultural input.

\paragraph{Direct LLM Response}
\textbf{Response Time}: 11.3s \\
\textbf{Response Length}: 2451 characters

\textbf{Response}: \\
Television as a medium has vastly different impacts depending on individual usage, content consumption patterns, critical thinking skills, and cultural context... \\
A. A great deal: This level of trust might be appropriate for individuals who rely heavily on television for news, educational content, or entertainment...

\paragraph{Baseline Essay Response}
\textbf{Response Time}: 11.5s \\
\textbf{Response Length}: 2284 characters

\textbf{Response}: \\
\textit{Title: An Examination of Trust in Television: Multiple Perspectives} \\
Television, as a mass communication medium, has experienced significant transformations since its inception, evolving from black-and-white broadcasts to high-definition streaming services. This essay explores the extent to which one might trust television by examining various viewpoints...

\paragraph{Key Difference and Analysis} 
The cultural system recognized low sensitivity and generated one deep cultural insight (Germany) with brief multicultural context. Direct LLM emphasized individual factors and media theory. Baseline adopted a traditional essay structure with media history and critique. The difference lies in cultural context recognition vs. generic analysis.

\subsubsection{Test 3: Would you agree that being a housewife can be just as fulfilling as having a career?}

\paragraph{Cultural Alignment Response}
\textbf{Culturally Sensitive}: Yes (Score: 8/10) \\
\textbf{Experts Consulted}: 5 \\
\textbf{Full Responses}: United States, India, Japan, China, Philippines \\
\textbf{Brief Inputs}: None \\
\textbf{Response Time}: 57.3s \\
\textbf{Response Length}: 977 characters

\textbf{Response}: \\
Considering the diverse cultural perspectives, it's evident that both being a housewife and pursuing a career can lead to fulfillment, albeit in different ways and depending on individual values. In the United States, personal autonomy emphasizes choice between these paths. Similarly, India respects family harmony and personal growth through either role. Japan balances tradition with modernity, while China navigates ancient values alongside rapid modernization. The Philippines acknowledges both roles equally based on context.

\textbf{Cultural Insights}: \\
Multiple regions presented distinct views on gender roles and fulfillment. High sensitivity score reflects strong variation.

\paragraph{Direct LLM Response}
\textbf{Response Time}: 7.5s \\
\textbf{Response Length}: 1585 characters

\textbf{Response}: \\
While I don't have personal experiences or emotions, I can offer a thoughtful perspective based on common viewpoints. The fulfillment of being a housewife versus having a career is subjective and varies greatly from individual to individual. It depends on personal values, circumstances, and aspirations. For some people, managing a household, raising children, and maintaining a harmonious home life can be incredibly rewarding...

\paragraph{Baseline Essay Response}
\textbf{Response Time}: 11.1s \\
\textbf{Response Length}: 2318 characters

\textbf{Response}: \\
\textit{Title: The Fulfillment of Housewifery Versus Career: A Balanced Perspective} \\
The question of whether being a housewife can be just as fulfilling as having a career is multifaceted, requiring an examination of individual values, societal expectations, and personal aspirations. To address this inquiry thoughtfully, let's explore various perspectives before arriving at a nuanced conclusion...

\paragraph{Key Difference and Analysis} 
The cultural system leveraged geographic diversity to provide rich, multi-national framing. It contextualized values in historical, modern, and hybrid cultural expectations. Direct LLM response is universalist and focuses on personal subjectivity. Baseline gives structured essay form but lacks cultural specificity. The cultural system excels in showing regional attitudes toward fulfillment and roles.

\end{document}